\documentclass[runningheads]{llncs}
\usepackage{graphicx}
\usepackage{comment}
\usepackage{amsmath,amssymb} 
\usepackage{color}
\usepackage{pifont}
\usepackage{multirow}
\usepackage{bbm}
\usepackage{subfig}
\usepackage{tablefootnote}
\usepackage{wrapfig,lipsum,booktabs}

\newcommand\samethanks[1][\value{footnote}]{\footnotemark[#1]}
\newcommand{\cmark}{\ding{51}}
\newcommand{\etal}{\textit{et al.}}
\newcommand{\stdev}[1]{{\scriptsize ${\pm {#1}}$}}

\usepackage{amsmath,amsfonts,bm}

\def\Figref#1{Figure~\ref{#1}}

\def\eqref#1{equation~\ref{#1}}
\def\Eqref#1{Equation~\ref{#1}}

\def\1{\bm{1}}

\newcommand{\test}{\mathcal{D_{\mathrm{test}}}}

\def\vc{{\bm{c}}}

\def\vh{{\bm{h}}}

\def\vu{{\bm{u}}}
\def\vv{{\bm{v}}}

\def\vx{{\bm{x}}}

\def\vz{{\bm{z}}}

\def\eva{{a}}

\def\mH{{\bm{H}}}

\def\mV{{\bm{V}}}
\def\mW{{\bm{W}}}
\def\mX{{\bm{X}}}

\DeclareMathAlphabet{\mathsfit}{\encodingdefault}{\sfdefault}{m}{sl}
\SetMathAlphabet{\mathsfit}{bold}{\encodingdefault}{\sfdefault}{bx}{n}

\begin{document}
\pagestyle{headings}
\mainmatter
\def\ECCVSubNumber{4629}  

\title{Environment-agnostic Multitask Learning for Natural Language Grounded Navigation} 

\titlerunning{Environment-agnostic Multitask Navigation}
%
\author{Xin Eric Wang\inst{1,2}\thanks{Equal contribution.} \thanks{Work done at Google.} \and
Vihan Jain\inst{3}\samethanks[1] \and
Eugene Ie\inst{3} \and 
William Yang Wang\inst{2} \and 
Zornitsa Kozareva\inst{3} \and
Sujith Ravi\inst{3,4}\samethanks[2]
} 

\authorrunning{X. Wang et al.}
%
\institute{
University of California, Santa Cruz \and
University of California, Santa Barbara \and
Google Research \and
Amazon }
\maketitle

\begin{abstract}
Recent research efforts enable study for natural language grounded navigation in photo-realistic environments, e.g., following natural language instructions or dialog. 
However, existing methods tend to overfit training data in seen environments and fail to generalize well in previously unseen environments. 
To close the gap between seen and unseen environments, we aim at learning a generalized navigation model from two novel perspectives:
(1) we introduce a multitask navigation model that can be seamlessly trained on both Vision-Language Navigation (VLN) and Navigation from Dialog History (NDH) tasks, which benefits from richer natural language guidance and effectively transfers knowledge across tasks;
(2) we propose to learn environment-agnostic representations for the navigation policy that are invariant among the environments seen during training, thus generalizing better on unseen environments.
Extensive experiments show that environment-agnostic multitask learning significantly reduces the performance gap between seen and unseen environments, and the navigation agent trained so outperforms baselines on unseen environments by 16\% (relative measure on success rate) on VLN and 120\% (goal progress) on NDH. Our submission to the CVDN leaderboard establishes a new state-of-the-art for the NDH task on the holdout test set. 
Code is available at \url{https://github.com/google-research/valan}.

\end{abstract}

\section{Introduction}
Navigation in visual environments by following natural language guidance~\cite{hemachandra2015learning} is a fundamental capability of intelligent robots that simulate human behaviors, because humans can easily reason about the language guidance and navigate efficiently by interacting with the visual environments. Recent efforts~\cite{anderson2018vision,das2018eqa,thomason:corl19:vdn,qi2020reverie} empower large-scale learning of natural language grounded navigation that is situated in photo-realistic simulation environments. 

Nevertheless, the generalization problem commonly exists for these tasks, especially indoor navigation: the agent usually performs poorly on unknown environments that have never been seen during training.
One of the leading causes of such behavior is data scarcity, as it is expensive and time-consuming to extend either visual environments or natural language guidance. 
The number of scanned houses for indoor navigation is limited due to high expense and privacy concerns. Besides, unlike vision-only navigation tasks~\cite{Mirowski:2018:StreetLearn,Mirowski2016LearningTN,xiazamirhe2018gibsonenv,habitat19iccv,ai2thor,li2020unsupervised} where episodes can be exhaustively sampled in simulation, natural language grounded navigation is supported by human demonstrated interaction in natural language. It is impractical to fully collect all the samples for individual tasks. 

Therefore, it is essential though challenging to efficiently learn a more generalized policy for natural language grounded navigation tasks from existing data~\cite{wu2018building,wu2018learning}.  
In this paper, we study how to resolve the generalization and data scarcity issues from two different angles.
First, previous methods are trained for one task at the time, so each new task requires training a new agent instance from scratch that can only
solve the one task on which it was trained. In this work, we propose a generalized multitask model for natural language grounded navigation tasks such as Vision-Language Navigation (VLN) and Navigation from Dialog History (NDH), aiming to efficiently transfer knowledge across tasks and effectively solve all the tasks simultaneously with one agent. 

Furthermore, even though there are thousands of trajectories paired with language guidance, the underlying house scans are restricted. For instance, the popular Matterport3D environment~\cite{Matterport3D} contains only 61 unique house scans in the training set. The current models perform much better in seen environments by taking advantage of the knowledge of specific houses they have acquired over multiple task completions during training, but fail to generalize to houses not seen during training.
To overcome this shortcoming, we propose an environment-agnostic learning method to learn a visual representation that is invariant to specific environments but can still support navigation. Endowed with the learned environment-agnostic representations, the agent is further prevented from the overfitting issue and generalizes better on unseen environments.

To the best of our knowledge, we are the first to introduce natural language grounded multitask and environment-agnostic training regimes and validate their effectiveness on VLN and NDH tasks. 
Extensive experiments demonstrate that our environment-agnostic multitask navigation model can not only efficiently execute different language guidance in indoor environments but also outperform the single-task baseline models by a large margin on both tasks. 
Besides, the performance gap between seen and unseen environments is significantly reduced.
Furthermore, our leaderboard submission for the NDH task establishes a new state-of-the-art outperforming the existing best agent by more than 66\% on the primary metric of goal progress on the holdout test set.
\section{Background}
\begin{figure}[t]
    \centering
    \includegraphics[clip, trim=0cm 6.5cm 0cm 0cm, width=\textwidth]{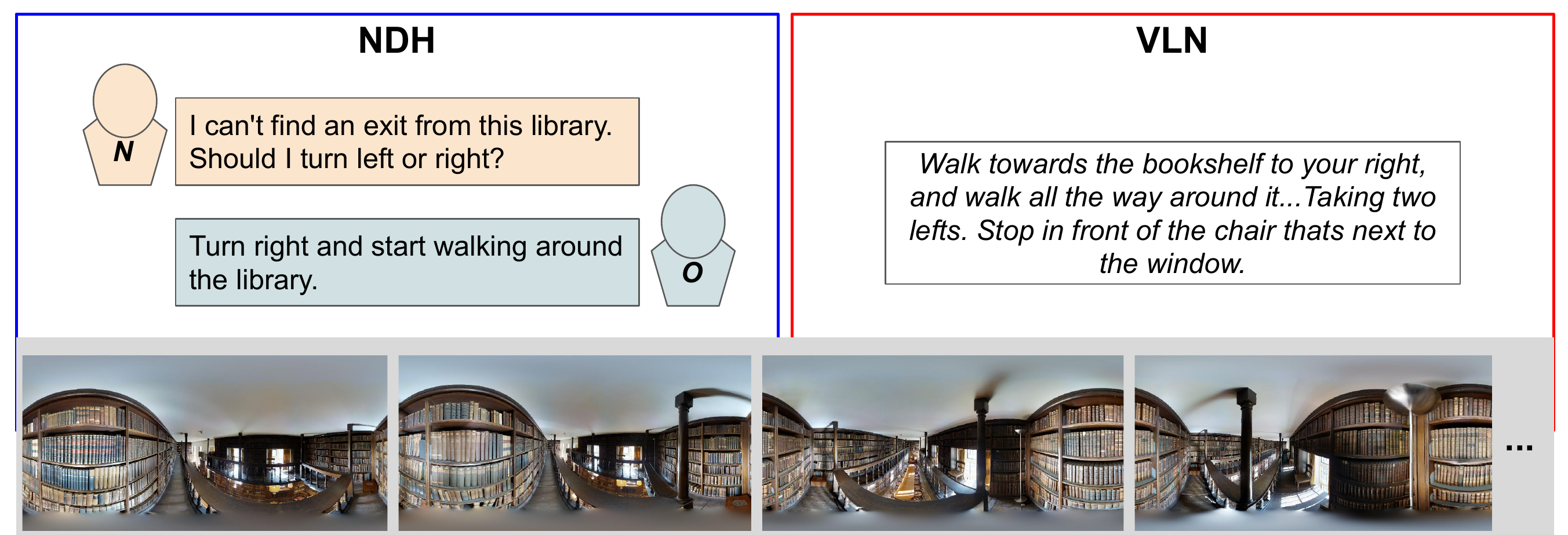}
    \caption{While the NDH task (left) requires an agent to navigate using dialog history between two human players - a navigator (N) who is trying to find the goal room with the help of an oracle (O), the VLN task (right) requires navigating using instructions written by human annotators.}
    \label{fig:task-depiction}
\end{figure}

\noindent\textbf{Vision-and-Language Navigation.~}
As depicted in \Figref{fig:task-depiction}, Vision-and-Language Navigation~\cite{anderson2018vision,ChenTouchdown2018} task requires an embodied agent to navigate in photo-realistic environments to carry out natural language instructions. For a given path, the associated natural language instructions describe the step-by-step guidance from the starting position to the target position.
The agent is spawned at an initial pose $p_0=(v_0, \phi_0, \theta_0)$, which includes the spatial location, heading and elevation angles. Given a natural language instruction $X = \{x_1, x_2, ..., x_n\}$, the agent is expected to perform a sequence of actions $\{a_1, a_2, ..., a_T\}$ and arrive at the target position $v_{tar}$ specified by the language instruction $X$.
In this work, we consider the VLN task defined for Room-to-Room (R2R)~\cite{anderson2018vision} dataset, which contains instruction-trajectory pairs across 90 different indoor environments (houses). 
The instructions for a given trajectory in the dataset on an average contain 29 words.
Previous VLN methods have studied various aspects to improve the navigation performance, such as planning~\cite{wang2018look}, data augmentation~\cite{fried2018speaker,tan-etal-2019-learning,fu2019counterfactual}, cross-modal alignment~\cite{Wang:2018:RCM,Huang2019TransferableRL}, progress estimation~\cite{ma2019self}, error correction~\cite{ma2019regretful,ke2019tactical}, interactive language assistance~\cite{nguyen2019vision,nguyen2019help} etc. This work tackles VLN via multitask learning and environment-agnostic learning, which is orthogonal to all these prior arts.

\noindent\textbf{Navigation from Dialog History.~}
Different from Visual Dialog~\cite{visdial} that involves dialog grounded in a single image, the recently introduced Cooperative Vision-and-Dialog Navigation (CVDN) dataset~\cite{thomason:corl19:vdn} includes interactive language assistance for indoor navigation, which consists of over 2,000 embodied, human-human dialogs situated in photo-realistic home environments. The task of Navigation from Dialog History (NDH) demonstrated in \Figref{fig:task-depiction}, is defined as: given a target object $t_0$ and a dialog history between humans cooperating to perform the task, the embodied agent must infer navigation actions towards the goal room that contains the target object. The dialog history is denoted as $<t_0, Q_1, A_1, Q_2, A_2, ..., Q_i, A_i>$, including the target object $t_0$, the questions $Q$ and answers $A$ till the turn $i$ ($0 \leq i \leq k$, where $k$ is the total number of Q-A turns from the beginning to the goal room). The agent, located in $p_0$, is trying to move closer to the goal room by inferring from the dialog history that happened before. The dialog for a given trajectory lasts 6 utterances (3 question-answer exchanges) and is 82 words long on an average.

\noindent\textbf{Multitask Learning.~}
The basis of multitask learning is the notion that tasks can serve as mutual sources of inductive bias for each other~\cite{Caruana93multitasklearning}. When multiple tasks are trained jointly, multitask learning causes the learner to prefer the hypothesis that explains all the tasks simultaneously, leading to more generalized solutions. Multitask learning has been successful in natural language processing~\cite{Collobert:2008:UAN:1390156.1390177}, speech recognition~\cite{speech:2013:Deng}, computer vision~\cite{rcnn:2015:Girshick}, drug discovery~\cite{Ramsundar2015MassivelyMN}, and Atari games~\cite{teh2017distral}. The deep reinforcement learning methods that have become very popular for training models on natural language grounded navigation tasks~\cite{Wang:2018:RCM,Huang:2019:Discriminator,Huang2019TransferableRL,tan-etal-2019-learning} are known to be data inefficient. In this work, we introduce multitask reinforcement learning for such tasks to improve data efficiency by positive transfer across related tasks. 

\noindent\textbf{Agnostic Learning.~}
A few studies on agnostic learning have been proposed recently. For example, Model-Agnostic
Meta-Learning (MAML)~\cite{finn2017maml} aims to train a model on a variety of learning tasks and solve a new task using only a few training examples. 
Liu \etal~\cite{liu2018unified} proposes a unified feature disentangler that learns domain-invariant representation across multiple domains for image translation. 
Other domain-agnostic techniques are also proposed for supervised~\cite{Li2018WhatsIA} and unsupervised domain adaption~\cite{romijnders2019domain,icml/PengHSS19}.
In this work, we pair the environment classifier with a gradient reversal layer~\cite{Ganin:2015:UDA} to learn an environment-agnostic representation that can be better generalized on unseen environments in a zero-shot fashion where no adaptation is involved.  

\noindent\textbf{Distributed Actor-Learner Navigation Learning Framework.~}
\label{subsec:dali}
To train models for the various language grounded navigation tasks like VLN and NDH, we use the VALAN framework~\cite{Lansing2019VALAN}, a distributed actor-learner learning infrastructure. The framework is inspired by IMPALA~\cite{pmlr-v80-espeholt18a} and uses its off-policy correction method called V-trace to scale reinforcement learning methods to thousands of machines efficiently. The framework additionally supports a variety of supervision strategies essential for navigation tasks such as teacher-forcing~\cite{anderson2018vision}, student-forcing~\cite{anderson2018vision} and mixed supervision~\cite{thomason:corl19:vdn}. The framework is built using TensorFlow~\cite{abadi2016tensorflow} and supports ML accelerators (GPU, TPU).
\section{Environment-agnostic Multitask Learning}

\begin{figure}[t]
    \centering
    \includegraphics[width=\textwidth]{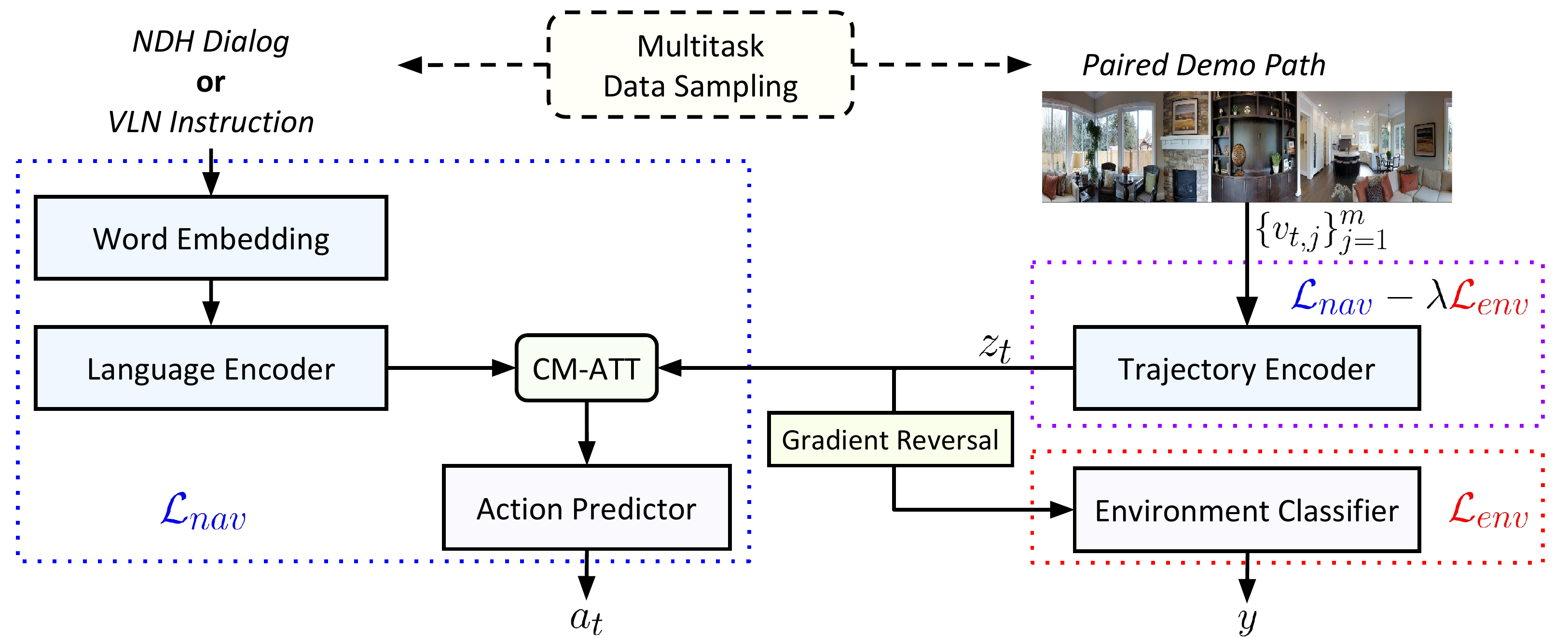}
    \caption{Overview of environment-agnostic multitask learning.}
    \label{fig:model}
\end{figure}

\subsection{Overview}
\label{subsec:overview}
Our environment-agnostic multitask navigation model is illustrated in \Figref{fig:model}. First, we adapt the reinforced cross-modal matching (RCM) model~\cite{Wang:2018:RCM} and make it seamlessly transfer across tasks by sharing all the learnable parameters for both NDH and VLN, including joint word embedding layer, language encoder, trajectory encoder, cross-modal attention module (CM-ATT), and action predictor.
Furthermore, to learn the environment-agnostic representation $\vz_t$, we equip the navigation model with an environment classifier whose objective is to predict which house the agent is. However, note that between trajectory encoder and environment classifier, a gradient reversal layer~\cite{Ganin:2015:UDA} is introduced to reverse the gradients back-propagated to the trajectory encoder, making it learn representations that are environment-agnostic and thus more generalizable in unseen environments.
During training, the environment classifier is minimizing the environment classification loss $\mathcal{L}_{env}$, while the trajectory encoder is maximizing $\mathcal{L}_{env}$ and minimizing the navigation loss $\mathcal{L}_{nav}$. The other modules are optimized with the navigation loss $\mathcal{L}_{nav}$ simultaneously.
Below we introduce multitask reinforcement learning and environment-agnostic representation learning. 
A more detailed model architecture is presented in Sec.~\ref{subsec:model-arch}.

\subsection{Multitask Reinforcement Learning}
In this section, we describe how we adapted the RCM agent model to learn the two tasks of VLN and NDH simultaneously. It is worth noting that even though both the VLN and NDH tasks use the same Matterport3D indoor environments~\cite{Matterport3D}, there are significant differences in the motivations and the overall objectives of the two tasks. While the natural language descriptions associated with the paths in the VLN task are step-by-step instructions to follow the ground-truth paths, the descriptions of the paths in the NDH task are series of question-answer interactions (dialog) between two human players which need not necessarily align sequentially with the ground-truth paths. This difference in the style of the two tasks also manifests in their respective datasets --- the average path description length and average path length in the NDH task's dataset are roughly three times that of the VLN task's dataset.
Furthermore, while the objective in VLN is to find the exact goal node in the environment (i.e., point navigation), the objective in NDH is to find the goal room that contains the specified object (i.e., room navigation).

\paragraph{Interleaved Multitask Data Sampling.}
To avoid overfitting individual tasks, we adopt an interleaved multitask data sampling strategy to train the model. Particularly, each data sample within a mini-batch can be from either task, so that the VLN instruction-trajectory pairs and NDH dialog-trajectory pairs are interleaved in a mini-batch though they may have different learning objectives. 

\paragraph{Reward Shaping.}
\label{subsec:reward_shaping}
Following prior art~\cite{wang2018look,Wang:2018:RCM}, we first implement a discounted cumulative reward function $R$ for the VLN and NDH tasks:
\begin{equation}
    R(s_t, a_t) = \sum_{t'=t}^T \gamma^{t'-t} r(s_{t'}, a_{t'})
\end{equation}
where $\gamma$ is the discounted factor. For the VLN task, we choose the immediate reward function such that the agent is rewarded at each step for getting closer to (or penalized for getting further from) the target location. At the end of the episode, the agent receives a reward only if it terminated successfully. Formally,

\begin{equation}
\label{eq:vln_reward}
r(s_{t'}, a_{t'}) = 
\begin{cases}
    d(s_{t'}, v_{tar}) - d(s_{t'+1}, v_{tar}) \quad \text{if } t' < T \\
    \mathbbm{1}[d(s_T, v_{tar}) \leq d_{th}] \qquad\qquad \text{if } t'=T
\end{cases}
\end{equation}
where $d(s_t, v_{tar})$ is the distance between state $s_t$ and the target location $v_{tar}$, $\mathbbm{1}[.]$ is the indicator function and $d_{th}$ is the maximum distance from $v_{tar}$ that the agent is allowed to terminate for success.

Different from VLN, the NDH task is essentially room navigation instead of point navigation because the agent is expected to reach a room that contains the target object. Suppose the goal room is occupied by a set of nodes $\{v_i\}_1^N$, we replace the distance function $d(s_t, v_{tar})$ in \Eqref{eq:vln_reward} with the minimum distance to the goal room $d_{room}(s_t, \{v_i\}_1^N)$ for NDH:
\begin{equation}
d_{room}(s_t, \{v_i\}_1^N) = \min_{1 \leq i \leq N} d(s_t, v_i)
\end{equation}

\paragraph{Navigation Loss.} Since human demonstrations are available for both VLN and NDH tasks, we use behavior cloning to constrain the learning algorithm to model state-action spaces that are most relevant to each task. Following previous works~\cite{Wang:2018:RCM}, we also use reinforcement learning to aid the agent's ability to recover from erroneous actions in unseen environments. During navigation model training, we adopt a mixed training strategy of reinforcement learning and behavior cloning, so the navigation loss function is:
\begin{equation}
    \mathcal{L}_{nav} = - \mathbb{E}_{a_t \sim \pi}[R(s_t, a_t) - b] - \mathbb{E}[\log \pi(a_t^*|s_t)]
\end{equation}
where we use REINFORCE policy gradients~\cite{Williams:1992:PolicyGradient} and supervised learning gradients to update the policy $\pi$. $b$ is the estimated baseline to reduce the variance and $a_t^*$ is the human demonstrated action.

\subsection{Environment-agnostic Representation Learning}
To further improve the navigation policy's generalizability, we propose to learn a latent environment-agnostic representation that is invariant among seen environments. The objective is to not learn the intricate environment-specific features that are irrelevant to general navigation (e.g. unique house appearances), preventing the model from overfitting to specific seen environments.
We can reformulate the navigation policy as 
\begin{equation}
    \pi(a_t | s_t) = p(a_t|\vz_t, s_t)p(\vz_t | s_t)
\end{equation}
where $\vz_t$ is a latent representation. 

As shown in \Figref{fig:model}, $p(a_t|\vz_t,s_t)$ is modeled by the policy module (including CM-ATT and action predictor) and $p(\vz_t|s_t)$ is modeled by the trajectory encoder.
In order to learn the environment-agnostic representation, we employ an environment classifier and a gradient reversal layer~\cite{Ganin:2015:UDA}. The environment classifier is parameterized to predict the house identity, so its loss function $\mathcal{L}_{env}$ is defined as
\begin{equation}
    \mathcal{L}_{env} = - \mathbb{E}[\log p(y=y^*|\vz_t)]
\end{equation}
where $y^*$ is the ground-truth house label. The gradient reversal layer has no parameters. It acts as an identity transform during forward-propagation, but multiplies the gradient by $-\lambda$ and passes it to
the trajectory encoder during back-propagation. Therefore, in addition to minimizing the navigation loss $\mathcal{L}_{nav}$, the trajectory encoder is also maximizing the environment classification loss $\mathcal{L}_{env}$. While the environment classifier is minimizing the classification loss conditioned on the latent representation $\vz_t$, the trajectory encoder is trying to increase the classifier's entropy, resulting in an adversarial learning objective.
\section{Model Architecture}
\label{subsec:model-arch}
\noindent\textbf{Language Encoder.~} 
The natural language guidance (instruction or dialog) is tokenized and embedded into n-dimensional space $\mX=\{\vx_1, \vx_2, ..., \vx_n\}$ where the word vectors $\vx_i$ are initialized randomly. The vocabulary is restricted to tokens that occur at least five times in the training instructions (the vocabulary used when jointly training VLN and NDH tasks is the union of the two tasks' vocabularies.). All out-of-vocabulary tokens are mapped to a single out-of-vocabulary identifier. The token sequence is encoded using a bi-directional LSTM~\cite{Schuster1997BidirectionalRN} to create $\mH^{\mX}$ following:
\begin{align}
    \mH^{{\mX}} &= [\vh_1^{\mX}; \vh_2^{\mX}; ...; \vh_n^{\mX}], 
    \quad \vh_t^{\mX} = \sigma(\overrightarrow{\vh}_t^{\mX}, \overleftarrow{\vh}_t^{\mX})  \\
    \overrightarrow{\vh}_t^{\mX} &= LSTM(\vx_t, \overrightarrow{\vh}_{t-1}^{\mX}),
    \quad \overleftarrow{\vh}_t^{\mX} = LSTM(\vx_t, \overleftarrow{\vh}_{t+1}^{\mX})
\end{align}
\noindent
where $\overrightarrow{\vh}_t^{\mX}$ and $\overleftarrow{\vh}_t^{\mX}$ are the hidden states of the forward and backward LSTM layers at time step $t$ respectively, and the $\sigma$ function is used to combine $\overrightarrow{\vh}_t^{\mX}$ and $\overleftarrow{\vh}_t^{\mX}$ into $\vh_t^{\mX}$.

\noindent\textbf{Trajectory Encoder.~}
Similar to benchmark models~\cite{fried2018speaker,Wang:2018:RCM,Huang2019TransferableRL}, at each time step $t$, the agent perceives a 360-degree panoramic view at its current location. The view is discretized into $k$ view angles ($k=36$ in our implementation, 3 elevations by 12 headings at 30-degree intervals).
The image at view angle $i$, heading angle $\phi$ and elevation angle $\theta$ is represented by a concatenation of the pre-trained CNN image features with the 4-dimensional orientation feature [sin $\phi$; cos $\phi$; sin $\theta$; cos $\theta$] to form $\vv_{t,i}$. 
The visual input sequence $\mV=\{\vv_1, \vv_2, ..., \vv_m\}$ is encoded using a LSTM to create $\mH^{\mV}$ following:
\begin{equation}
    \mH^{\mV} = [\vh_1^{\mV}; \vh_2^{\mV}; ...; \vh_m^{\mV}], \quad \text{where~} \vh_t^{\mV} = LSTM(\vv_t, \vh_{t-1}^{\mV})
\end{equation}
$\vv_t=\text{Attention}(\vh_{t-1}^{\mV}, \vv_{t,1..k})$ is the attention-pooled representation of all view angles using previous agent state $\vh_{t-1}$ as the query. We use the dot-product attention~\cite{vaswani2017attention} hereafter.

\noindent\textbf{Policy Module.~}
The policy module comprises of cross-modal attention (CM-ATT) unit as well as an action predictor. The agent learns a policy $\pi_{\theta}$ over parameters $\theta$ that maps the natural language instruction $\mX$ and the initial visual scene $\vv_1$ to a sequence of actions $[\eva_1, \eva_2, ..., \eva_n]$. The action space which is common to VLN and NDH tasks consists of navigable directions from the current location. The available actions at time $t$ are denoted as $\vu_{t,1..l}$, where $\vu_{t,j}$ is the representation of the navigable direction $j$ from the current location obtained similarly to $\vv_{t,i}$. The number of available actions, $l$, varies per location, since graph node connectivity varies. Following Wang \etal~\cite{Wang:2018:RCM}, the model predicts the probability $p_d$ of each navigable direction $d$ using a bilinear dot product:
\begin{equation}
p_d = \text{softmax}([\vh_t^{\mV}; \vc_t^{\text{text}}; \vc_t^{\text{visual}}]\mW_c(\vu_{t,d}\mW_u)^T)
\end{equation}
where $\vc_t^{\text{text}} = \text{Attention}(\vh_t^{\mV}, \vh^{\mX}_{1..n})$ and $\vc_t^{\text{visual}} = \text{Attention}(\vc_t^{\text{text}}, \vv_{t, 1..k})$. $\mW_c$ and $\mW_u$ are learnable parameters.

\noindent\textbf{Environment Classifier.~} 
The environment classifier is a two-layer perceptron with a SoftMax layer as the last layer. Given the latent representation $\vz_t$ (which is $\vh_t^{\mV}$ in our setting), the classifier generates a probability distribution over the house labels.

\section{Experiments}
\label{sec:experiments}

\subsection{Experimental Setup}
\label{subsec:exp-setup}
\noindent\textbf{Implementation Details.} 
We use a 2-layer bi-directional LSTM for the instruction encoder, where the size of LSTM cells is 256 in each direction. The inputs to the encoder are 300-dimensional embeddings initialized randomly. For the visual encoder, we use a 2-layer LSTM with a cell size of 512. The encoder inputs are image features derived as mentioned in Sec.~\ref{subsec:model-arch}. The cross-modal attention layer size is 128 units. The environment classifier has one hidden layer of size 128 units, followed by an output layer of size equal to the number of classes. The negative gradient multiplier $\lambda$ in the gradient reversal layer is empirically tuned and fixed at a value of $1.3$ for all experiments. During training, some episodes in the batch are identical to available human demonstrations in the training dataset, where the objective is to increase the agent's likelihood of choosing human actions (behavioral cloning~\cite{Bain:1999:Cloning}). The rest of the episodes are constructed by sampling from the agent's own policy. For the NDH task, we deploy mixed supervision similar to Thomason \etal~\cite{thomason:corl19:vdn}, where the navigator's or oracle's path is selected as ground-truth depending on if the navigator was successful in reaching the correct end node following the question-answer exchange with the oracle or not.
In the experiments, unless otherwise stated, we use the entire dialog history from the NDH task for model training. 
\emph{All the reported results in subsequent studies are averages of at least three independent runs.}

\noindent\textbf{Evaluation Metrics.~} 
The agents are evaluated on two datasets, namely \textit{Validation Seen} that contains new paths from the training environments and \textit{Validation Unseen} that contains paths from previously unseen environments. The evaluation metrics for VLN task are as follows: 
\textit{Path Length (PL)} measures the total length of the predicted path; 
\textit{Navigation Error (NE)} measures the distance between the last nodes in the predicted and the reference paths; 
\textit{Success Rate (SR)} measures how often the last node in the predicted path is within some threshold distance of the last node in the reference path; 
\textit{Success weighted by Path Length (SPL)}~\cite{Anderson:2018:Evaluation} measures Success Rate weighted by the normalized Path Length; 
and \textit{Coverage weighted by Length Score (CLS)}~\cite{Jain2019StayOT} measures predicted path's conformity to the reference path weighted by length score. 
For the NDH task, the agent's progress is defined as a reduction (in meters) from the distance to the goal region at the agent's first position versus at its last position~\cite{thomason:corl19:vdn}.

\subsection{Environment-agnostic Multitask Learning}
\label{subsec:exp-envagnostic-multitask}
\begin{figure}
    \centering
    \includegraphics[width=\textwidth]{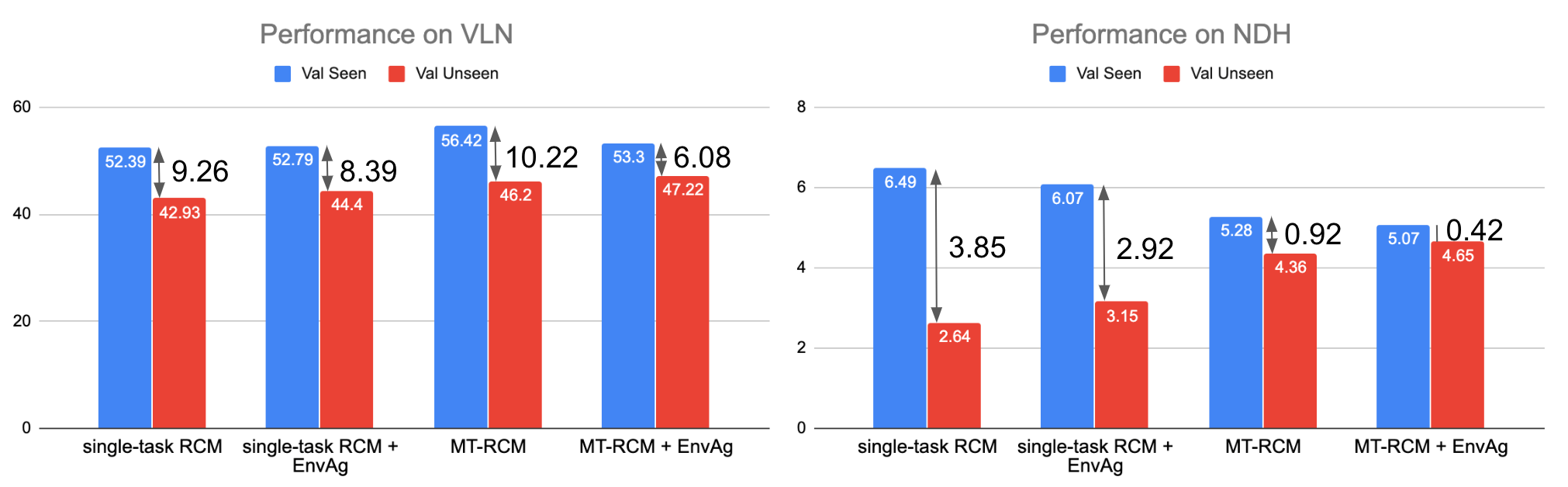}
    \caption{Visualizing performance gap between seen and unseen environments for VLN (success rate) and NDH (progress) tasks.}
    \label{fig:perf_gap_viz}
\end{figure}

Table~\ref{tab:summary} shows the results of training the navigation model using environment-agnostic learning (\textit{EnvAg}) as well as multitask learning (\textit{MT-RCM}). First, both learning methods independently help the agent learn more generalized navigation policy, as is evidenced by a significant reduction in agent's performance gap between seen and unseen environments (better visualized with \Figref{fig:perf_gap_viz}). For instance, the performance gap in goal progress on the NDH task drops from 3.85m to 0.92m using multitask learning, and the performance gap in success rate on the VLN task drops from 9.26\% to 8.39\% using environment-agnostic learning. 
Second, the two techniques are complementary---the agent's performance when trained with both the techniques simultaneously improves on unseen environments compared to when trained separately. Finally, we note here that \textit{MT-RCM $+$ EnvAg} outperforms the baseline goal progress of 2.10m~\cite{thomason:corl19:vdn} on NDH validation unseen dataset by more than 120\%. At the same time, it outperforms the equivalent RCM baseline~\cite{Wang:2018:RCM} of 40.6\% success rate by more than 16\% (relative measure) on VLN validation unseen dataset.

\begin{table}[t]
    \caption{The agent's performance under different training strategies. The single-task RCM (ST-RCM) model is independently trained and tested on VLN or NDH tasks. The standard deviation across 3 independent runs is reported.}
    \label{tab:summary}
    \centering
    \setlength\tabcolsep{2.0pt}
\resizebox{\textwidth}{!}{
    \begin{tabular}{clcccccc}
    \toprule
    \multirow{3}{*}{\textbf{Fold}} & \multirow{3}{*}{\textbf{Model}}  & \textbf{NDH}  & \multicolumn{5}{c}{\textbf{VLN}}  \\
                            \cmidrule(lr){3-3} \cmidrule(lr){4-8}
    &                        & Progress $\uparrow$ & PL & NE $\downarrow$ & SR $\uparrow$ & SPL $\uparrow$ & CLS $\uparrow$  \\
    \midrule
    \multirow{7}{*}{\rotatebox[origin=c]{90}{Val Seen}} 
    & seq2seq~\cite{thomason:corl19:vdn} & 5.92 &       &      &       &       &          \\
    & RCM~\cite{Wang:2018:RCM}\tablefootnote{The equivalent RCM model without intrinsic reward is used as the benchmark.}  &      & 12.08 & \textbf{3.25} & \textbf{67.60} & - & -  \\
    \cmidrule(lr){2-8}
    & \textbf{Ours} \\
    & ST-RCM            & \textbf{6.49} \stdev{0.95}     & 10.75 \stdev{0.26}  & 5.09 \stdev{0.49}  & 52.39 \stdev{3.58}  & 48.86 \stdev{3.66}  & 63.91 \stdev{2.41} \\
    & ST-RCM $+$ EnvAg  &  6.07 \stdev{0.56}    & 11.31 \stdev{0.26} &    4.93 \stdev{0.49}  & 52.79 \stdev{3.72} & 48.85 \stdev{3.71} & 63.26 \stdev{2.31} \\
    & MT-RCM             & 5.28 \stdev{0.56} & 10.63 \stdev{0.10} & 5.09 \stdev{0.05} & \textbf{56.42} \stdev{1.21} & \textbf{49.67} \stdev{1.07} & \textbf{68.28} \stdev{0.16} \\
    & MT-RCM $+$ EnvAg   & 5.07 \stdev{0.45} & 11.60 \stdev{0.30} & \textbf{4.83} \stdev{0.12} & 53.30 \stdev{0.71} & 49.39 \stdev{0.74} & 64.10 \stdev{0.16} \\
    \midrule
    \multirow{7}{*}{\rotatebox[origin=c]{90}{Val Unseen}} 
    & seq2seq~\cite{thomason:corl19:vdn} & 2.10 &       &      &       &       &   \\
    & RCM~\cite{Wang:2018:RCM}  &      & 15.00 & 6.02 & 40.60 & - & - \\
    \cmidrule(lr){2-8}
    & \textbf{Ours} \\

    & ST-RCM            &  2.64 \stdev{0.06}  & 10.60 \stdev{0.27} & 6.10 \stdev{0.06} & 42.93 \stdev{0.21} & 38.88 \stdev{0.20} & 54.86 \stdev{0.92} \\
    & ST-RCM $+$ EnvAg  &  3.15 \stdev{0.29}  & 11.36 \stdev{0.27}   & 5.79 \stdev{0.06} & 44.40 \stdev{2.14} & 40.30 \stdev{2.12} & 55.77 \stdev{1.31} \\
    & MT-RCM            & 4.36 \stdev{0.17} & 10.23 \stdev{0.14} & \textbf{5.31} \stdev{0.18} & 46.20 \stdev{0.55} & \textbf{44.19} \stdev{0.64} & 54.99 \stdev{0.87}  \\
    & MT-RCM $+$ EnvAg  & \textbf{4.65} \stdev{0.20} & 12.05 \stdev{0.23} & 5.41 \stdev{0.20} & \textbf{47.22} \stdev{1.00} & 41.80 \stdev{1.11} & \textbf{56.22} \stdev{0.87} \\
    \bottomrule
    \end{tabular}
}
\end{table}

To further validate our results on NDH task, we evaluated the \textit{MT-RCM $+$ EnvAg} agent on the test set of NDH dataset which is held out as the CVDN challenge\footnote{\url{https://evalai.cloudcv.org/web/challenges/challenge-page/463/leaderboard/1292}}. Table~\ref{tab:ndh-test-submission} shows that our submission to the leaderboard with \textit{MT-RCM $+$ EnvAg} establishes a new state-of-the-art on this task outperforming the existing best agent by more than 66\%.

\begingroup
\setlength{\tabcolsep}{10pt} 
\begin{table}
    \caption{Comparison on CVDN Leaderboard Test Set. Note that the metric \emph{Progress} is the same as \emph{dist\_to\_end\_reduction}.}
    \label{tab:ndh-test-submission}
    \centering
    \begin{tabular}{clc}
    \toprule
    &  Agent  &  Progress $\uparrow$     \\
      \midrule
      \multirow{2}{*}{Baselines} & Random   & 0.83 \\
                                 & Shortest Path Agent (\textit{upper bound})  & 9.76  \\
      \midrule
      \multirow{2}{*}{Leaderboard Submissions} & Seq2Seq~\cite{thomason:corl19:vdn} & 2.35  \\
                                 & \textbf{MT-RCM $+$ EnvAg}  & \textbf{3.91} \\
    \bottomrule
    \end{tabular}
    
\end{table}
\endgroup

\subsection{Multitask Learning}
\label{subsec:exp-multitask}

\begingroup
\setlength{\tabcolsep}{1.25pt} 
\begin{table}[t]
\small
  \caption{Comparison of agent performance when trained separately \textit{vs.} jointly on VLN and NDH tasks.}
  \label{tab:multi-task}
  \centering
  \begin{tabular}{llcccccccccc}
    \toprule
                   &                          & \multicolumn{5}{c}{\textbf{NDH Evaluation}}  & \multicolumn{5}{c}{\textbf{VLN Evaluation}} \\
                                              \cmidrule(lr){3-7} \cmidrule(lr){8-12}
    \textbf{Fold}  & \textbf{Model} & \multicolumn{4}{c}{\textbf{Inputs for NDH}} & \textbf{Progress} & \textbf{PL} & \textbf{NE} & \textbf{SR} & \textbf{SPL} & \textbf{CLS} \\
                   &                          & $t_o$ & $A_i$ & $Q_i$ & $A_{1:i-1};Q_{1:i-1}$           & $\uparrow$ & & $\downarrow$ & $\uparrow$ & $\uparrow$ & $\uparrow$ \\
    \midrule
    \multirow{8}{*}{\parbox[c]{1.1cm}{Val\\Seen}} & \multirow{4}{*}{NDH-RCM}   & \cmark & & & & \textbf{6.97} \\
                               &                            & \cmark & \cmark & & & \textbf{6.92} \\
                               &                            & \cmark & \cmark & \cmark & & \textbf{6.47} \\
                               &                            & \cmark & \cmark & \cmark & \cmark & \textbf{6.49} \\
                                                            \cmidrule(lr){2-7}
                               & VLN-RCM                   & & & & & & 10.75 & 5.09 & 52.39 & 48.86 & 63.91  \\
                               \cmidrule(lr){2-12}
                               & \multirow{4}{*}{MT-RCM}   & \cmark & & & & 3.00 & 11.73 & 4.87 & 54.56 & 52.00 & 65.64 \\
                               &                            & \cmark & \cmark & & & 5.92 & 11.12 & 4.62 & 54.89 & \textbf{52.62} & 66.05 \\
                               &                            & \cmark & \cmark & \cmark & & 5.43 & 10.94 & \textbf{4.59} & 54.23 & 52.06 & 66.93 \\
                               &                            & \cmark & \cmark & \cmark & \cmark & 5.28 & 10.63 & 5.09 & \textbf{56.42} & 49.67 & \textbf{68.28} \\
    \midrule
    \multirow{8}{*}{\parbox[c]{1.1cm}{Val\\Unseen}} & \multirow{4}{*}{NDH-RCM}   & \cmark & & & & 1.25 \\
                               &                            & \cmark & \cmark & & & 2.69 \\
                               &                            & \cmark & \cmark & \cmark & & 2.69 \\
                               &                            & \cmark & \cmark & \cmark & \cmark & 2.64 \\
                                                            \cmidrule(lr){2-7}
                               & VLN-RCM                   & & & & & & 10.60 & 6.10 & 42.93 & 38.88 & 54.86 \\
                               \cmidrule(lr){2-12}
                               & \multirow{4}{*}{MT-RCM}   & \cmark & & & & \textbf{1.69} & 13.12 & 5.84 & 42.75 & 38.71 & 53.09 \\
                               &                            & \cmark & \cmark & & & \textbf{4.01} & 11.06 & 5.88 & 42.98 & 40.62 & 54.30 \\
                               &                            & \cmark & \cmark & \cmark & & \textbf{3.75} & 11.08 & 5.70 & 44.50 & 39.67 & 54.95 \\
                               &                            & \cmark & \cmark & \cmark & \cmark & \textbf{4.36} & 10.23 & \textbf{5.31} & \textbf{46.20} & \textbf{44.19} & \textbf{54.99} \\
    \bottomrule
  \end{tabular}
\end{table}
\endgroup

We then conduct studies to examine cross-task transfer using multitask learning alone.  
First, we experiment multitasking learning with access to different parts of the dialog---the target object $t_o$, the last oracle answer $A_i$, the prefacing navigator question $Q_i$, and the full dialog history.
Table~\ref{tab:multi-task} shows the results of jointly training \textit{MT-RCM} model on VLN and NDH tasks. 
(1) \emph{ Does VLN complement NDH?} Yes, consistently. On NDH Val Unseen, \textit{MT-RCM} consistently benefits from following shorter paths with step-by-step instructions in VLN for all kinds of dialog inputs. It shows that VLN can serve as an essential task to boost learning of primitive action-and-instruction following and therefore support more complicated navigation tasks like NDH.
(2) \emph{Does NDH complement VLN?} Yes, under certain conditions. From the results on VLN Val Unseen, we can observe that MT-RCM with only target objects as the guidance performs equivalently or slightly worse than \textit{VLN-RCM}, showing that extending visual paths alone (even with final targets) is not helpful in VLN. 
But we can see a consistent and gradual increase in the success rate of \textit{MT-RCM} on the VLN task as it is trained on paths with richer dialog history from the NDH task. This shows that the agent benefits from more fine-grained information about the path implying the importance given by the agent to the language instructions in the task.
(3) Multitask learning improves the generalizability of navigation models: the seen-unseen performance gap is narrowed.
(4) As a side effect, results of different dialog inputs on NDH Val Seen \emph{versus} Unseen verify the essence of language guidance in generalizing navigation to unseen environments.


\begin{figure}[!h]
    \centering
    \includegraphics[width=\textwidth]{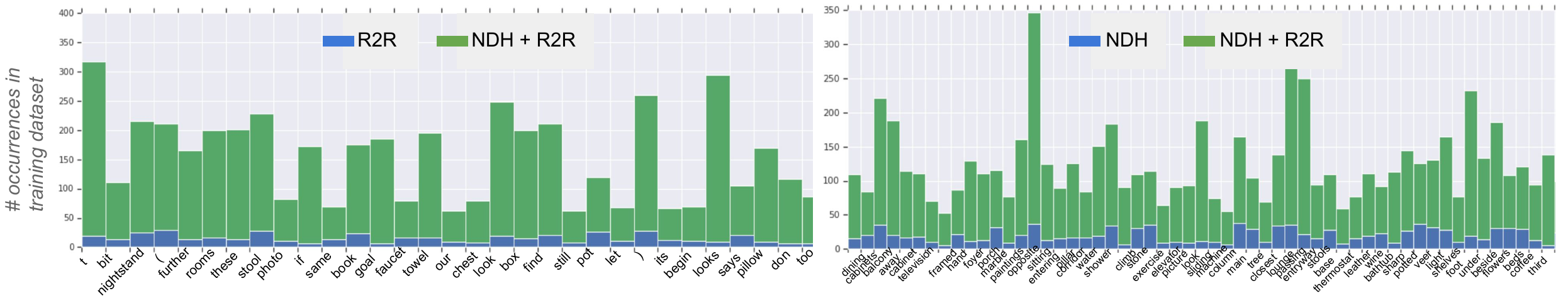}
    \caption{Selected tokens from the vocabulary for VLN (left) and NDH (right) tasks which gained more than 40 additional occurrences in the training dataset due to joint-training.}
    \label{fig:word-freq}
\end{figure}

\begin{table}[!h]
\small
    \caption{Comparison of agent performance when language instructions are encoded by separate \textit{vs.} shared encoder for VLN and NDH tasks.}
    \label{tab:encoder-ablation}
    \centering
\resizebox{\textwidth}{!}{
    \begin{tabular}{lcccccccccccc}
    \toprule
    \multirow{4}{*}{\textbf{Language Encoder}}  & \multicolumn{6}{c}{\textbf{Val Seen}}  &  \multicolumn{6}{c}{\textbf{Val Unseen}} \\
                            \cmidrule(lr){2-7} \cmidrule(lr){8-13}
                            & \textbf{NDH}  & \multicolumn{5}{c}{\textbf{VLN}}  & \textbf{NDH}  & \multicolumn{5}{c}{\textbf{VLN}} \\
                            \cmidrule(lr){2-2} \cmidrule(lr){3-7} \cmidrule(lr){8-8} \cmidrule(lr){9-13}
                            & Progress $\uparrow$ & PL & NE $\downarrow$ & SR $\uparrow$ & SPL $\uparrow$ & CLS $\uparrow$ & Progress $\uparrow$ & PL & NE $\downarrow$ & SR $\uparrow$ & SPL $\uparrow$ & CLS $\uparrow$ \\
    \midrule
    Shared   & \textbf{5.28} & 10.63 & 5.09 & \textbf{56.42} & \textbf{49.67} & \textbf{68.28} & \textbf{4.36} & 10.23 & \textbf{5.31} & \textbf{46.20} &   \textbf{44.19} & \textbf{54.99} \\
    Separate & 5.17 & 11.26 & \textbf{5.02} & 52.38 & 48.80 & 64.19 & 4.07 & 11.72 & 6.04 & 43.64 & 39.49 & 54.57 \\
    \bottomrule
    \end{tabular}
}
\end{table}

Besides, we show multitask learning results in better language grounding through more appearance of individual words in \Figref{fig:word-freq} and shared semantic encoding of the whole sentences in Table~\ref{tab:encoder-ablation}.
\Figref{fig:word-freq} illustrates that under-represented tokens in each of the individual tasks get a significant boost in the number of training samples.
Table~\ref{tab:encoder-ablation} shows that the model with shared language encoder for NDH and VLN tasks outperforms the model that has separate language encoders for the two tasks, hence demonstrating the importance of parameter sharing during multitask learning. 

Furthermore, we observed that the agent's performance improves significantly when trained on a mixture of VLN and NDH paths even when the size of the training dataset is fixed, advancing the argument that multitask learning on NDH and VLN tasks complements the agent's learning. More details of the ablation studies can be found in the Appendix.

\begin{figure}[t]
    \centering
    \includegraphics[clip, trim=0cm 7.4cm 0cm 0cm, width=\textwidth]{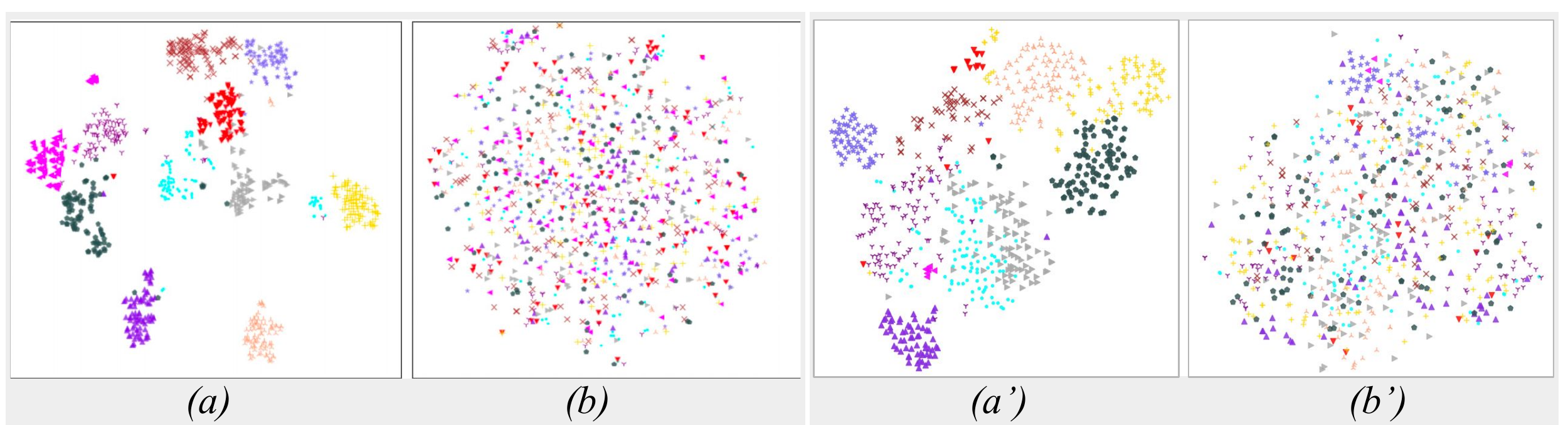}
    \caption{t-SNE visualization of trajectory encoder's output for VLN task across 11 different color-coded seen (\emph{a},\emph{b}) and unseen (\emph{a'},\emph{b'}) environments. The depicted representations in \emph{(a)} and \emph{(a')} are learned with environment-aware objective while those in \emph{(b)} and \emph{(b')} are learned with environment-agnostic objective.}
    \label{fig:t_sne}
\end{figure}

\subsection{Environment-agnostic Learning}
\label{subsec:exp-envagnostic}
From Table~\ref{tab:summary}, it can be seen that both VLN and NDH tasks benefit from environment-agnostic learning independently. 
To further examine the generalization property of environment-agnostic learning, we train a model with the opposite objective---learn to correctly predict the navigation environments by removing the gradient reversal layer (\emph{environment-aware learning}).
The results in Table~\ref{tab:env-agnostic} demonstrate that environment-aware learning leads to overfitting on the training dataset as the performance on environments seen during training consistently increases for both the tasks. In contrast, environment-agnostic learning leads to a more generalized navigation policy that performs better on unseen environments. 
\Figref{fig:t_sne} further shows that due to environment-aware learning, the model learns to represent visual inputs from the same environment closer to each other while the representations of different environments are farther from each other resulting in a clustering learning effect. On the other hand, environment-agnostic learning leads to more general representation across different environments, which results in better performance on unseen environments.

\begin{table}[t]
\small
    \caption{Environment-agnostic \textit{versus} environment-aware learning.}
    \label{tab:env-agnostic}
    \centering
    \setlength\tabcolsep{1pt}
    \subfloat[Comparison on NDH.]{
    \resizebox{0.32\textwidth}{!}{  
    \begin{tabular}{l cc}
        \toprule
         \multirow{3}{*}{\textbf{Model}} & \textbf{Val Seen}  & \textbf{Val Unseen} \\
        \cmidrule(lr){2-2} \cmidrule(lr){3-3}
        & Progress $\uparrow$ & Progress $\uparrow$ \\
        \midrule
        RCM & 6.49 & 2.64 \\
        EnvAware & \textbf{8.38} & 1.81\\
        EnvAg & 6.07 & \textbf{3.15} \\
        \bottomrule
    \end{tabular}
    }}
    ~
    \subfloat[Comparison on VLN.]{
    \resizebox{0.64\textwidth}{!}{  
    \begin{tabular}{l ccccc ccccc}
        \toprule
        \multirow{3}{*}{\textbf{Model}}  & \multicolumn{5}{c}{\textbf{Val Seen}}  &  \multicolumn{5}{c}{\textbf{Val Unseen}} \\
                                \cmidrule(lr){2-6} \cmidrule(lr){7-11}
                                 & PL & NE $\downarrow$ & SR $\uparrow$ & SPL $\uparrow$ & CLS $\uparrow$ & PL & NE $\downarrow$ & SR $\uparrow$ & SPL $\uparrow$ & CLS $\uparrow$ \\
        \midrule
        RCM            & 10.75 & 5.09 & 52.39 & 48.86 & 63.91  & 10.60 & 6.10 & 42.93 & 38.88 & 54.86 \\
        EnvAware       & 10.30 & \textbf{4.36} & \textbf{57.59}  & \textbf{54.05}  &    \textbf{68.49}   & 10.13 & 6.30 & 38.83  & 35.65  & 54.79 \\
        EnvAg  & 11.31 &    4.93       & 52.79           & 48.85            & 63.26  & 11.36    & \textbf{5.79} & \textbf{44.40} & \textbf{40.30} & \textbf{55.77} \\
        \bottomrule
    \end{tabular}
    }}
\end{table}

\subsection{Reward Shaping for NDH task}
\label{subsec:exp-reward-shaping-short}
As discussed in Sec.~\ref{subsec:reward_shaping}, we conducted studies to shape the reward for the NDH task. Table~\ref{tab:reward-shaping} presents the results of training the agent with access to different parts of the dialog history. The results demonstrate that the agents rewarded for getting closer to the goal room consistently outperform the agents rewarded for getting closer to the exact goal location. This proves that using a reward function better aligned with the NDH task's objective yields better performance than other reward functions.

\begingroup
\setlength{\tabcolsep}{1.5pt} 
\begin{table}[t]
\small
  \caption{Average agent progress towards goal room when trained using different rewards and mixed supervision strategy.}
  \label{tab:reward-shaping}
  \centering
  \begin{tabular}{cclcccccc}
    \toprule
    & & \multirow{2}{*}{\textbf{Model}} & \multicolumn{4}{c}{\textbf{Inputs}}         & \multicolumn{2}{c}{\textbf{Goal Progress (m)}} \\
          & &       & $t_0$ & $A_i$ & $Q_i$ & $A_{1:i-1};Q_{1:i-1}$ & Val Seen & Val Unseen \\
    \midrule
    & \multirow{6}{*}{Baselines} & Shortest-Path Agent & & & & & 9.52 & 9.58 \\
                               &                            & Random Agent & & & & & 0.42 &1.09 \\
                                                            \cmidrule(lr){3-9}
                               &                            & \multirow{4}{*}{Seq2Seq~\cite{thomason:corl19:vdn}} & \cmark & & & & 5.71 & \textbf{1.29} \\
                               &                            &                            & \cmark & \cmark & & & 6.04 & 2.05 \\
                               &                            &                            & \cmark & \cmark & \cmark & & 6.16 & 1.83 \\
                               &                            &                            & \cmark & \cmark & \cmark & \cmark & 5.92 & 2.10 \\
                               \cmidrule(lr){1-9}
                               & \multirow{8}{*}{Ours}      & \multirow{4}{*}{\parbox[t]{4cm}{NDH-RCM\\(distance to goal location)}} & \cmark & & & & 4.18 & 0.42 \\
                               &                            &                                                  & \cmark & \cmark & & & 4.96 & 2.34 \\
                               &                            &                                                  & \cmark & \cmark & \cmark & & 4.60 & 2.25 \\
                               &                            &                                                  & \cmark & \cmark & \cmark & \cmark & 5.02 & 2.58 \\
                                                            \cmidrule(lr){3-9}
                               &                            & \multirow{4}{*}{\parbox[t]{4cm}{NDH-RCM\\(distance to goal room)}}     & \cmark & & & & \textbf{6.97} & 1.25 \\
                               &                            &                                                  & \cmark & \cmark & & & \textbf{6.92} & \textbf{2.69} \\
                               &                            &                                                  & \cmark & \cmark & \cmark & & \textbf{6.47} & \textbf{2.69} \\
                               &                            &                                                  & \cmark & \cmark & \cmark & \cmark & \textbf{6.49} & \textbf{2.64}\\
    \bottomrule
  \end{tabular}
\end{table}
\endgroup

\section{Conclusion}
In this work, we presented an environment-agnostic multitask learning framework to learn generalized policies for agents tasked with natural language grounded navigation.
We applied the framework to train agents that can simultaneously solve two popular and challenging tasks in the space: Vision-and-Language Navigation and Navigation from Dialog History.
We showed that our approach effectively transfers knowledge across tasks and learns more generalized environment representations.
As a result, the trained agents not only close down the performance gap between seen and unseen environments but also outperform the single-task baselines on both tasks by a significant margin.
Furthermore, the studies show the two approaches of multitask learning and environment-agnostic learning independently benefit the agent learning and complement each other. 
There are possible future extensions to our work---\textit{MT-RCM} can further be adapted to other language-grounded navigation datasets (e.g., Touchdown~\cite{ChenTouchdown2018}, TalkTheWalk~\cite{de2018talk}, StreetLearn~\cite{Mirowski:2018:StreetLearn}); and complementary techniques like environmental dropout~\cite{tan-etal-2019-learning} can be combined with environment-agnostic learning to learn more general representations.
\section{Acknowledgements}
We thank Larry Lansing for his enormous efforts in scaling the VALAN framework which made it possible to run experiments at scale.

%
%
\bibliographystyle{splncs04}
\bibliography{main.bbl}

\newpage
\appendix

\section{Appendix}

\subsection{Multitask Learning with Total Training Paths Fixed}

To verify whether multitask learning helps only due to implicit data augmentation which increases the number of training paths for both the tasks, we conducted an experiment by fixing the total number of training paths. We fix the size of training dataset to be exactly $4,742$ paths (which is the same as the number of paths in the NDH task's training dataset) and replace a fraction of the paths by the paths from the VLN task's training dataset.  The results in Table~\ref{tab:mt_fixed_train_data} show that the agent's performance on previously unseen environments in NDH task improves significantly when trained jointly on NDH paths mixed with a small fraction of VLN paths. Since the total training paths are fixed, there is no benefit due to data augmentation which furthers the argument that multitask learning on NDH and VLN tasks complements the agent's learning. As expected, the agent's performance on NDH task degrades when trained on datasets containing smaller fractions of NDH paths but larger fractions of VLN paths.

\begin{table}
    \caption{Comparison of agent's performance on NDH task when trained on fixed number of paths. The paths belong to either of the two tasks to support multitask learning.}
    \label{tab:mt_fixed_train_data}
    \centering
    \setlength\tabcolsep{3.0pt}
    \begin{tabular}{lccccccc}
    \toprule
                            &  \multicolumn{7}{c}{Fraction of VLN paths (\%)} \\
                            \cmidrule(lr){2-8}
                            &   0     &  10   &   20    &  30   &  40   &  60   &  80  \\
    \midrule
    Progress (Val Seen)     &   6.49  & 6.21  &   5.69  &  5.72 &  5.82 &  5.74 & 3.66 \\
    Progress (Val Unseen)   &   2.64  & 3.13  &   3.09  &  3.31 &  2.80 &  2.86 & 2.48 \\
    \bottomrule
    \end{tabular}
\end{table}

\subsection{Detailed Ablation on Parameter Sharing of Language Encoder}
Table~\ref{tab:text-encoder-ablation} presents a more detailed analysis from Table~\ref{tab:encoder-ablation} with access to different parts of dialog history. The models with shared language encoder consistently outperform those with separate encoders.

\begingroup
\setlength{\tabcolsep}{1.1pt} 
\begin{table}
\small
  \caption{Comparison of agent's performance when language instructions are encoded by separate \textit{vs.} shared encoder for VLN and NDH tasks.}
  \label{tab:text-encoder-ablation}
  \centering
  \begin{tabular}{llcccccccccc}
  \toprule
                   &                          & \multicolumn{5}{c}{\textbf{NDH Evaluation}}  & \multicolumn{5}{c}{\textbf{VLN Evaluation}} \\
                                              \cmidrule(lr){3-7} \cmidrule(lr){8-12}
    \textbf{Fold}  & \textbf{\parbox[c]{1.75cm}{Language\\Encoder}} & \multicolumn{4}{c}{\textbf{Inputs for NDH}} & \textbf{Progress} & \textbf{PL} & \textbf{NE} & \textbf{SR} & \textbf{SPL} & \textbf{CLS} \\
                   &                          & $t_0$ & $A_i$ & $Q_i$ & $A_{1:i-1};Q_{1:i-1}$           & $\uparrow$ & & $\downarrow$ & $\uparrow$ & $\uparrow$ & $\uparrow$ \\
    \midrule
    \multirow{8}{*}{\parbox[c]{1.2cm}{Val\\Seen}} & \multirow{4}{*}{Shared} & \cmark & & & & \textbf{3.00} & 11.73 & 4.87 & 54.56 & 52.00 & 65.64 \\
                              &              & \cmark & \cmark & & & \textbf{5.92} & 11.12 & \textbf{4.62} & 54.89 & \textbf{52.62} & 66.05 \\
                              &              & \cmark & \cmark & \cmark & & \textbf{5.43} & 10.94 & 4.59 & 54.23 & 52.06 & 66.93 \\
                              &              & \cmark & \cmark & \cmark & \cmark & \textbf{5.28} & 10.63 & 5.09 & \textbf{56.42} & 49.67 & \textbf{68.28}  \\
                              \cmidrule{2-12}
                              & \multirow{4}{*}{Separate} & \cmark & & & & 2.85 & 11.43 & 4.81 & 54.66 & 51.11 & 65.37 \\
                              &                            & \cmark & \cmark & & & 4.90 & 11.92 & 4.92 & 53.64 & 49.79 & 61.49 \\
                              &                            & \cmark & \cmark & \cmark & & 5.07 & 11.34 & 4.76 & 55.34 & 51.59 & 65.52 \\
                              &                            & \cmark & \cmark & \cmark & \cmark & 5.17 & 11.26 & 5.02 & 52.38 & 48.80 & 64.19 \\
    \midrule
    \multirow{8}{*}{\parbox[c]{1.2cm}{Val\\Unseen}} & \multirow{4}{*}{Shared} & \cmark & & & & 1.69 & 13.12 & 5.84 & 42.75 & 38.71 & 53.09 \\
                              &              & \cmark & \cmark & & & \textbf{4.01} & 11.06 & 5.88 & 42.98 & 40.62 & 54.30 \\
                              &              & \cmark & \cmark & \cmark & & \textbf{3.75} & 11.08 & 5.70 & 44.50 & 39.67 & 54.95 \\
                              &              & \cmark & \cmark & \cmark & \cmark & \textbf{4.36} & 10.23 & \textbf{5.31} & \textbf{46.20} & \textbf{44.19} & \textbf{54.99}  \\
                              \cmidrule{2-12}
                              & \multirow{4}{*}{Separate} & \cmark & & & & \textbf{1.79} & 11.85 & 6.01 & 42.43 & 38.19 & 54.01 \\
                              &                            & \cmark & \cmark & & & 3.66 & 12.59 & 5.97 & 43.45 & 38.62 & 53.49 \\
                              &                            & \cmark & \cmark & \cmark & & 3.51 & 12.23 & 5.89 & 44.40 & 39.54 & 54.55 \\
                              &                            & \cmark & \cmark & \cmark & \cmark & 4.07 & 11.72 & 6.04 & 43.64 & 39.49 & 54.57 \\
  \bottomrule
  \end{tabular}
\end{table}
\endgroup

\subsection{VLN Leaderboard Submission}

Table~\ref{tab:vln_leaderboard} shows the performance of our \textit{MT-RCM $+$ EnvAg} agent on the test set of the R2R dataset for the VLN task. Our \textit{MT-RCM $+$ EnvAg} agent outperforms the comparable baseline \textit{RCM} on the primary navigation metrics of SR and SPL which proves the effectiveness of multitask and environment-agnostic learning. It is worth noting that the baselines scoring high on the test set use additional techniques like data augmentation and pre-training which were not explored in this work but are complementary to our techniques of multitask learning and environment-agnostic learning.

\begin{table}[!h]
    \caption{VLN Leaderboard results for R2R test dataset.}
    \label{tab:vln_leaderboard}
    \centering
    \setlength\tabcolsep{4.0pt}
    \begin{tabular}{lcccc}
    \toprule
    \textbf{Model} & \textbf{PL} & \textbf{NE} $\downarrow$ & \textbf{SR} $\uparrow$ & \textbf{SPL} $\uparrow$ \\
    \midrule
    Human                 & 11.85 & 1.61 & 86 & 76 \\
    Random                & 9.93  & 9.77 & 13 & 12 \\
    \midrule
    Seq2Seq~\cite{anderson2018vision}               & 8.13  & 7.85 & 20 & 18 \\
    Look Before You Leap~\cite{wang2018look}  & 9.15  & 7.53 & 25 & 23 \\
    Speaker-Follower~\cite{fried2018speaker}      & 14.82 & 6.62 & 35 & 28 \\
    Self-Monitoring~\cite{ma2019self}       & 18.04 & 5.67 & 48 & 35 \\
    RCM~\cite{Wang:2018:RCM}                   & 11.97 & 6.12 & 43 & 38 \\
    The Regretful Agent~\cite{ma2019regretful}   & 13.69 & 5.69 & 48 & 40 \\
    ALTR~\cite{Huang2019TransferableRL}                  & 10.27 & 5.49 & 48 & 45 \\
    Press~\cite{li_2019_robust}                 & 10.77 & 5.49 & 49 & 45 \\
    EnvDrop~\cite{tan-etal-2019-learning}               & 11.66 & 5.23 & 51 & 47 \\
    PREVALENT~\cite{Hao_2020_CVPR}            & 10.51 & 5.30 & 54 & 51 \\
    \midrule
    \textbf{MT-RCM $+$ EnvAg} (Ours) & 13.35 & 6.03 & 45 & 40 \\
    \bottomrule
    \end{tabular}
\end{table}



\end{document}